\documentclass[11pt,a4paper]{article}

\usepackage[table]{xcolor}
\usepackage[utf8]{inputenc}
\usepackage[hyperref]{acl2017}
\usepackage{times}
\usepackage{latexsym}
\usepackage{CJKutf8}
\usepackage{amsmath}
\usepackage{graphicx}

\aclfinalcopy 

\definecolor{mauve}{RGB}{224 176 255}
\definecolor{Myorange}{cmyk}{0,0.42,1,0}
\definecolor{Myblue}{rgb}{0.8,0.85,1}
\definecolor{Mygrey}{gray}{0.75}

\definecolor{orange}{HTML}{E500CB}

\definecolor{one}{RGB}{214, 209, 255}  
\definecolor{two}{RGB}{211, 205, 254}  
\definecolor{three}{RGB}{208, 201, 254} 
\definecolor{four}{RGB}{205, 198, 254}  
\definecolor{five}{RGB}{203, 184, 254} 
\definecolor{six}{RGB}{200, 191, 254}  
\definecolor{seven}{RGB}{197, 187, 254} 
\definecolor{eight}{RGB}{195, 184, 254} 
\definecolor{nine}{RGB}{192, 180, 254} 
\definecolor{ten}{RGB}{189, 177, 254} 

\definecolor{eleven}{RGB}{189, 173, 254} 
\definecolor{twelve}{RGB}{186, 169, 254} 
\definecolor{thirteen}{RGB}{183, 165, 254} 
\definecolor{fourteen}{RGB}{180, 161, 254} 
\definecolor{fifteen}{RGB}{177, 157, 254} 
\definecolor{sixteen}{RGB}{173, 153, 254} 
\definecolor{seventeen}{RGB}{170, 149, 253} 
\definecolor{eighteen}{RGB}{167, 145, 253} 
\definecolor{nineteen}{RGB}{163, 141, 253} 
\definecolor{twenty}{RGB}{160, 137, 253} 

\definecolor{twone}{RGB}{157, 133, 253} 
\definecolor{twtwo}{RGB}{154, 129, 253} 
\definecolor{twthree}{RGB}{151, 125, 253} 
\definecolor{twfour}{RGB}{148, 121, 252} 
\definecolor{twfive}{RGB}{145, 117, 252} 
\definecolor{twsix}{RGB}{142, 113, 252} 
\definecolor{twseven}{RGB}{139, 110, 252} 
\definecolor{tweight}{RGB}{136, 106, 252} 
\definecolor{twnine}{RGB}{133, 102, 252} 
\definecolor{thirty}{RGB}{130, 98, 252} 

\definecolor{thone}{RGB}{127, 94, 252} 
\definecolor{thtwo}{RGB}{124, 90, 252} 
\definecolor{ththree}{RGB}{121, 86, 251} 
\definecolor{thfour}{RGB}{118, 82, 251} 
\definecolor{thfive}{RGB}{115, 78, 251} 
\definecolor{thsix}{RGB}{112, 74, 251} 
\definecolor{thseven}{RGB}{109, 70, 251} 
\definecolor{theight}{RGB}{106, 66, 251} 
\definecolor{thnine}{RGB}{103, 62, 251} 
\definecolor{fourty}{RGB}{100, 58, 251} 

\definecolor{foone}{RGB}{97, 54, 250} 
\definecolor{fotwo}{RGB}{94, 50, 250} 
\definecolor{fothree}{RGB}{91, 46, 250} 
\definecolor{fofour}{RGB}{88, 42, 250} 
\definecolor{fofive}{RGB}{85, 38, 250} 
\definecolor{fosix}{RGB}{82, 34, 250} 
\definecolor{foseven}{RGB}{79, 30, 250} 
\definecolor{foeight}{RGB}{76, 27, 250} 
\definecolor{fonine}{RGB}{73, 23, 250} 
\definecolor{fourty}{RGB}{70, 20, 250} 

\definecolor{fione}{RGB}{67, 17, 250} 
\definecolor{fitwo}{RGB}{64, 13, 250} 
\definecolor{fithree}{RGB}{61, 9, 250} 
\definecolor{fifour}{RGB}{58, 5, 250} 
\definecolor{fifive}{RGB}{55, 3, 250} 
\definecolor{fisix}{RGB}{52, 2, 250} 
\definecolor{fiseven}{RGB}{49, 1, 250} 

\usepackage{calc}

\newlength{\DepthReference}
\newlength{\HeightReference}
\newlength{\Width}%

\newcommand{\MyColorBox}[2][red]%
{%
    \settowidth{\Width}{#2}%
    \colorbox{#1}%
    {%
        \raisebox{-\DepthReference}%
        {%
                \parbox[b][\HeightReference+\DepthReference][c]{\Width}{\centering#2}%
        }%
    }%
}

\newcommand\blfootnote[1]{%
  \begingroup
  \renewcommand\thefootnote{}\footnote{#1}%
  \addtocounter{footnote}{-1}%
  \endgroup
}

\title{
A Recurrent Neural Model with Attention for \\ the Recognition of Chinese Implicit Discourse Relations}

\author{Samuel R\"onnqvist$^{1,2,\ast}$\textnormal{,} Niko Schenk$^{2,\ast}$ \textnormal{and} Christian Chiarcos$^{2}$\\ 
  $^{1}$Turku Centre for Computer Science -- TUCS, \AA bo Akademi University, Turku, Finland\\
	$^{2}$Applied Computational Linguistics Lab, Goethe University, Frankfurt am Main, Germany\\ 
  {\tt sronnqvi@abo.fi}\\ 
	{\tt \{schenk,chiarcos\}@informatik.uni-frankfurt.de}
}

\date{}

\begin{document}
\maketitle
\begin{abstract}
We introduce an attention-based Bi-LSTM for Chinese implicit discourse relations and demonstrate that modeling argument pairs as a joint sequence can outperform word order-agnostic approaches. 
Our model benefits from a partial sampling scheme and   
 is conceptually simple, yet achieves state-of-the-art performance on the Chinese Discourse Treebank.
We also visualize its attention activity to illustrate
the model's ability to selectively focus on the relevant parts of an input sequence.

\end{abstract}

\section{Introduction}

\noindent \blfootnote{$^{\ast}$Both first authors contributed equally to this work.}True text understanding is one of the key goals in Natural Language Processing and requires capabilities beyond the lexical semantics of individual words or phrases. Natural language descriptions are typically driven by an inter-sentential coherent structure, exhibiting specific \emph{discourse} properties, which in turn contribute significantly to the global meaning of a text. Automatically detecting how meaning units are organized benefits practical downstream applications, such as question answering \citep{sun2007discourse}, recognizing textual entailment \citep{Hickl:2008:UDC:1599081.1599124}, 
sentiment analysis \citep{DBLP:conf/naacl/TrivediE13},
or text summarization \citep{DBLP:conf/emnlp/HiraoYNYN13}.

Various formalisms in terms of semantic coherence frameworks have been proposed to account for these contextual assumptions \cite{mann88b,lascarides:asher:1993a,journals/cogsci/Webber04}. The annotation schemata of the Penn Discourse Treebank \cite[PDTB]{prasad08} and the Chinese Discourse Treebank \cite[CDTB]{citeulike:11019973}, for instance, define discourse units as syntactically motivated character spans in the text, augmented with relations pointing from the second argument (\emph{Arg2}, prototypically, a discourse unit associated with an explicit discourse marker) to its antecedent, i.e., the discourse unit \emph{Arg1}.
Relations are labeled with a relation type (its sense) and the associated discourse marker.
Both, PDTB and CDTB, distinguish
\emph{explicit} from \emph{implicit} relations depending on the presence of such a marker (e.g., \emph{because}/\begin{CJK*}{UTF8}{gbsn} 因 \end{CJK*}).\footnote{The set of relation types and senses is completed by alternative lexicalizations (\textsc{AltLex}/discourse marker rephrased), and entity relations (\textsc{EntRel}/anaphoric coherence).
}
Sense classification for implicit relations is by far more challenging because the argument pairs lack the marker as an important feature. Consider, for instance, the following example from the CDTB as implicit \textsc{Conjunction}:


\smallskip
\textbf{\emph{Arg1}}:
\begin{CJK*}{UTF8}{gbsn}
  会谈	就	一些	原则	和	具体	问题	进行	了	深入	讨论	，	达成	了	一些	谅解 
\textit{\ \	In the talks, they discussed some principles and specific questions in depth, and reached some understandings}
\end{CJK*}

\textbf{\emph{Arg2}}:
\begin{CJK*}{UTF8}{gbsn}
  双方	一致	认为	会谈	具有	积极	成果 
\textit{\ \ Both sides agree that the talks have positive results}
\end{CJK*}

\smallskip
\noindent \textbf{Motivation:}
Previous work on implicit sense labeling is heavily feature-rich and requires domain-specific, semantic lexicons \cite{Pitler:2009:ASP:1690219.1690241,Feng:2012:TDP:2390524.2390534,huang-chen:2011:IJCNLP-2011}. Only recently, resource-lean architectures have been proposed. These promising neural methods attempt to infer latent representations appropriate for implicit relation classification \cite{DBLP:conf/emnlp/ZhangSXLDY15,ji-haffari-eisenstein:2016:N16-1,DBLP:conf/acl/ChenZLQH16}. So far, unfortunately, these models have been evaluated \emph{only} on four top-level senses---sometimes even with inconsistent evaluation setups.\footnote{E.g., four binary classifiers vs. four-way classification.}
Furthermore, most systems have initially been designed for the English PDTB and involve complex, task-specific architectures \cite{DBLP:conf/emnlp/LiuL16}, while discourse modeling techniques for Chinese have received very little attention in the literature and are still seriously underrepresented in terms of publicly available systems. What is more, over 80\% of all words in Chinese discourse relations are implicit---compared to only 52\% in English \cite{citeulike:11019973}. 

Recently, in the context of the CoNLL 2016 shared task \cite{xue-EtAl:2016:CoNLL-ST}, a first independent evaluation platform beyond class level has been established. Surprisingly, the best performing neural architectures to date are standard \emph{feedforward} networks, cf. \newcite{K16-2004,K16-2005,K16-2010}. Even though these specific models 
completely ignore word order within arguments, such feedforward architectures have been claimed by \newcite{DBLP:journals/corr/RutherfordDX16} to generally outperform any thoroughly-tuned recurrent architecture.

\smallskip
\noindent \textbf{Our Contribution:}
In this work, we release the first attention-based \emph{recurrent} neural sense classifier, specifically developed for Chinese implicit discourse relations.
 Inspired by \newcite{DBLP:conf/acl/ZhouSTQLHX16}, our system is a practical adaptation of the recent advances in relation modeling
 extended by a novel sampling scheme.

Contrary to previous assertions by \newcite{DBLP:journals/corr/RutherfordDX16}, our model demonstrates superior performance over traditional bag-of-words approaches with feedfoward networks by treating discourse arguments as a joint sequence.
We evaluate our method within an independent framework and show that it performs very well beyond standard class-level predictions, achieving state-of-the-art accuracy on the CDTB test set.

We illustrate how our model's attention mechanism provides means to highlight those parts of an input sequence
that are relevant for the classification decision, and thus, it may enable a better understanding of the implicit
discourse parsing problem. Our proposed
network architecture is flexible and largely language-independent as it operates only
on word embeddings. 
It stands out due to its structural simplicity and builds a solid ground for further development towards other textual domains.


\section{Approach}

\begin{figure}%
\includegraphics[width=1.0\columnwidth]{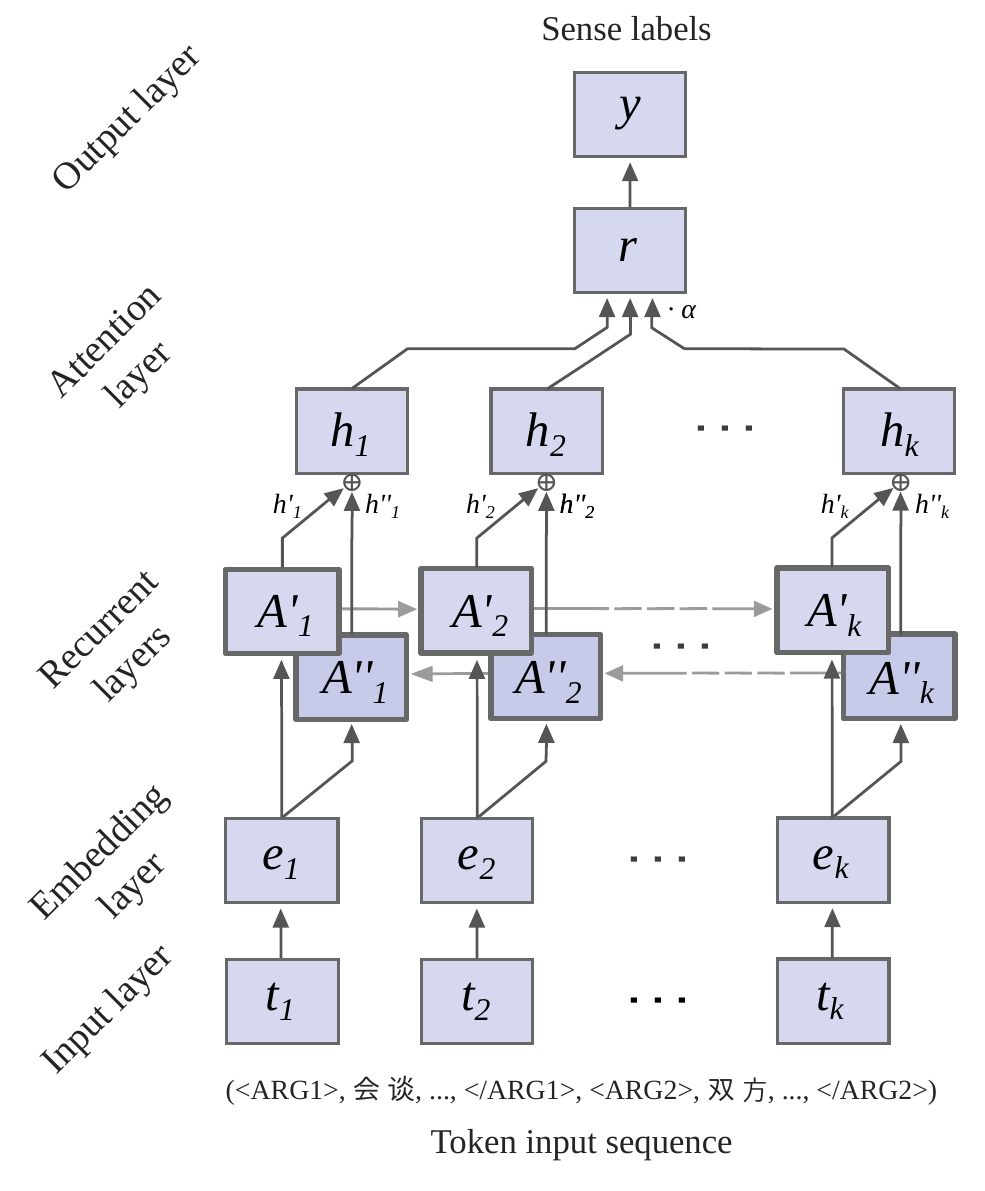}%
\caption{The attention-based bidirectional LSTM network for the task of modeling argument pairs for Chinese implicit discourse relations.}%
\label{networkarch}%
\end{figure}


We propose the use of an attention-based bidirectional Long Short-Term Memory \cite[LSTM]{Hochreiter:1997:LSM:1246443.1246450} network to predict senses of discourse relations. The model draws upon previous work on LSTM, in particular its bidirectional mode of operation \cite{DBLP:journals/nn/GravesS05}, attention mechanisms for recurrent models \cite{DBLP:journals/corr/BahdanauCB14,hermann2015teaching}, and the combined use of these techniques for entity relation recognition in annotated sequences \cite{DBLP:conf/acl/ZhouSTQLHX16}. 
More specifically, our model is a flexible recurrent neural network with capabilities to \emph{sequentially} inspect tokens and to highlight which parts of the input sequence are most informative for the discourse relation recognition task, using the weighting provided by the attention mechanism.
Furthermore, the model benefits from a novel sampling scheme for arguments, as elaborated below.
The system is learned in an end-to-end manner and consists of multiple layers, which are illustrated in Figure \ref{networkarch}.

First, token sequences are taken as input and special markers (\textless ARG1\textgreater, \textless/ARG1\textgreater , etc.) are inserted into the corresponding positions to inform the model on the start and end points of argument spans.
This way, we can ensure a general flexibility in modeling discourse units and could easily extend them with additional context, for instance. In our experiments on implicit arguments, only the tokens in the respective spans are considered.
Note that, unlike previous works, our approach models \emph{Arg1}-\emph{Arg2} pairs as a \emph{joint} sequence and does not first compute intermediate representations of arguments separately.

Second, an input layer encodes tokens using one-hot vector representations ($t_i$ for tokens at positions $i\in[1,k]$), and a subsequent embedding layer provides a dense representation ($e_i$) to serve as input for the recurrent layers. The embedding layer is initialized using pre-trained word vectors, in our case 300-dimensional Chinese Gigaword vectors \cite{chinese-gigaword}.\footnote{\url{http://www.cs.brandeis.edu/~clp/conll16st/dataset.html}}
These embeddings are further tuned as the network is trained towards the prediction task. Embeddings for unknown tokens, e.g., markers, are trained by backpropagation only. Note that, tokens, markers and the pre-trained vectors represent the only source of information for the prediction task.

For the recurrent setup, we use a layer of LSTM networks in a bidirectional manner, in order to better capture dependencies between parts of the input sequence by inspection of both left and right-hand-side contexts at each time step. The LSTM holds a state representation as a continuous vector passed to the subsequent time step, and it is capable of modeling long-range dependencies due to its gated memory.
The forward ($A'$) and backward ($A''$) LSTMs traverse the sequence $e_i$, producing sequences of vectors $h'_i$ and $h''_i$ respectively, which are then summed together (indicated by $\oplus$ in Figure \ref{networkarch}).

The resulting sequence of vectors $h_i$ is reduced into a single vector and fed to the final softmax output layer in order to classify the sense label $y$ of the discourse relation. This vector may be obtained either as the final vector $h$ produced by an LSTM, or through pooling of all $h_i$, or by using attention, i.e., as a weighted sum over $h_i$. While the model may be somewhat more difficult to optimize using attention, it provides the added benefit of interpretability, as the weights highlight to what extent the classifier considers the LSTM state vectors at each token during modeling. This is particularly interesting for discourse parsing, as most previous approaches have provided little support for pinpointing the driving features in each argument span.

Finally, the attention layer contains the trainable vector $w$ (of the same dimensionality as vectors $h_i$) which is used to dynamically produce a weight vector $\alpha$ over time steps $i$ by:

\[
\alpha = softmax(w^{T}tanh(H))
\]
where $H$ is a matrix consisting of vectors $h_i$. The output layer $r$ is the weighted sum of vectors in $H$:
\[
r = H\alpha^T
\]

\noindent
\textbf{Partial Argument Sampling:} For the purpose of enlarging the instance space of training items in the CDTB, and thus, in order to improve the predictive performance of the model, we propose a novel \emph{partial sampling} scheme of arguments, whereby the model is trained and validated on sequences containing both arguments, as well as \emph{single} arguments. A data point $(a_1,a_2,y)$, with $a_i$ being the token sequence of argument $i$, is expanded into $\{(a_1,a_2,y),(a_1,a_2,y),(a_1,y),(a_2,y)\}$. We duplicate bi-argument samples $(a_1,a_2,y)$ (in training and development data only) to balance their frequencies against single-argument samples. 

Two lines of motivation support the inclusion of single argument training examples, grounded in linguistics and machine learning, respectively. 
 First, it has been shown that single arguments in isolation can evoke a strong expectation towards a certain implicit discourse relation, cf. \newcite{asr2015uniform} and, in particular, \newcite{rohdehorton} in their psycholinguistic study on \emph{implicit causality verbs}. 
Second, the procedure may encourage the model to learn better representations of individual argument spans in support of  modeling of arguments in composition, cf. \newcite{lecun2015deep}.
Due to these aspects, we believe this data augmentation technique to be effective in reinforcing the overall robustness of our model.

\smallskip
\noindent{\textbf{Implementational Details:}} 
We train the model using fixed-length sequences of 256 tokens with zero padding at the beginning of shorter sequences and truncate longer ones. 
Each LSTM has a vector dimensionality of 300, matching the embedding size. The model is regularized by $0.5$ dropout rate between the layers and weight decay ($2.5e^{-6}$) on the LSTM inputs. 
We employ Adam optimization \cite{DBLP:journals/corr/KingmaB14} using the cross-entropy loss function with mini batch size of 80.\footnote{The model is implemented in \emph{Keras} \url{https://keras.io/}.} 

\begin{table*}[t!]
\begin{center}
\begin{tabular}{ c  l  c  r  c  l  c }
\noalign{\smallskip}
  \multicolumn{3}{c}{CDTB Development Set} &  & 
    \multicolumn{3}{c}{CDTB Test Set}  \\
  \cline{1-3}\cline{4-7}
  Rank & System & \% accuracy &  & Rank & System & \% accuracy\\
  \hline 
\rowcolor{gray!10}  1  & \footnotesize  \newcite{K16-2004} & 73.53 &\    & 1 & \footnotesize \newcite{K16-2004} & 72.42 \\
  2  & \footnotesize \newcite{K16-2010}  & 71.57  &\   & 2 & \footnotesize \newcite{K16-2005} & 71.87 \\
\rowcolor{gray!10}    3  & \footnotesize \newcite{K16-2005} & 70.59  &\    & 3 & \footnotesize \newcite{K16-2007} & 70.47 \\
  4  & \footnotesize   \newcite{K16-2007} & 68.30   &\    & 4 & \footnotesize \newcite{K16-2010} & 67.41 \\
\rowcolor{gray!10}  5  & \footnotesize  \newcite{K16-2006} & 66.67 &\   & 5 & \footnotesize \newcite{K16-2006} & 64.07 \\
  6  & \footnotesize \newcite{K16-2006} & 61.44 &\        & 6 & \footnotesize \newcite{K16-2006} & 63.51 \\
\rowcolor{gray!10}  7  & \footnotesize \newcite{K16-2022} & 21.90 &\     & 7 & \footnotesize \newcite{K16-2022} & 21.73 \\
\hline
       &  \textbf{This Paper:}  & \textbf{93.52$^{\ast}$} &\    &  &  \textbf{This Paper:} & \textbf{73.01} \\
\end{tabular}
\end{center}
\caption{Non-explicit parser scores on the official CoNLL 2016 CDTB development and test sets. ($^{\ast}$Scores on development set are obtained through partial sampling and are not directly comparable.)}
\label{rankings}
\end{table*}

\begin{table}%
\begin{tabular}{l*{3}{r}r}
Sense Label       & \small Training &\small  Dev't &\small  Test &  \\
\hline
\rowcolor{gray!10}\textsc{Conjunction} &  5,174 & 189  & 228  & \\
\rowcolor{gray!10} \small majority class    & \small (66.3\%)  & \small (62.8\%)  & \small (64.8\%) & \\
\textsc{Expansion}   & 1,188 & 48 & 40 &   \\
\rowcolor{gray!10}\textsc{EntRel}      & 1,099 & 50 & 71 &   \\
\textsc{Causation}   & 187 & 10 & 8 &   \\
\rowcolor{gray!10}\textsc{Contrast}    & 66 & 3 & 1 &   \\
\textsc{Purpose}     & 56 & 1 & 3 &   \\
\rowcolor{gray!10}\textsc{Conditional} & 26 & 0 & 1 &   \\
\textsc{Temporal}    & 26 & 0 & 0 &   \\
\rowcolor{gray!10}\textsc{Progression} & 7 & 0 & 0 &   \\
\hline
\# impl. rels & 7,804 & 301 & 352 &   \\
\end{tabular}
\caption{Implicit sense labels in the CDTB.}
\label{labeldist}
\end{table}



\section{Evaluation}

We evaluate our recurrent model on the CoNLL 2016 shared task data\footnote{\url{http://www.cs.brandeis.edu/~clp/conll16st/}} which include the official training, development and test sets of the CDTB; cf. Table \ref{labeldist} for an overview of the implicit sense distribution.\footnote{Note that, in the CDTB, implicit relations appear almost \emph{three times more often} than explicit relations. Out of these, 65\% appear within the same sentence. Finally, 25 relations in the training set have two labels.}
 
 In accordance with previous setups \cite{DBLP:journals/corr/RutherfordDX16}, we treat entity relations (\textsc{EntRel}) as implicit and exclude \textsc{AltLex} relations.
In the evaluation, we focus on the \emph{sense-only} track, the subtask for which gold arguments are provided and a system is supposed to label a given argument pair with the correct sense. 
The results are shown in Table \ref{rankings}. 

With our proposed architecture it is possible to correctly label 257/352 (73.01\%) of implicit relations on the test set, outperforming the best feedforward system of \newcite{K16-2004} and all other word order-agnostic approaches. Development and test set performances suggest the robustness of our approach and its ability to generalize to unseen data. 

\smallskip
\noindent \textbf{Ablation Study:} 
We perform an ablation study to quantitatively assess the contribution of two of the characteristic aspects of our model. First, we compare the use of the attention mechanism against the simpler alternative of feeding the final LSTM hidden vectors ($h'_{k}$ and $h''_{1}$) directly to the output layer. When attention is turned off, this yields an absolute decrease in performance of 2.70\% on the test set, which is substantial and significant according to a Welch two-sample t-test ($p$ $<$ .001). Second, we independently compare the use of the partial sampling scheme against training on the standard argument pairs in the CDTB. Here, the absence of the partial sampling scheme yields an absolute decrease in accuracy of 5.74\% ($p$ $<$ .001), which demonstrates its importance for achieving competitive performance on the task.


\smallskip
\noindent \textbf{Performance on the PDTB:} As a side experiment, we investigate the model's language independence by applying it to the implicit argument pairs of the English PDTB. Due to computational time constraints we do not optimize hyperparameters, but instead train the model using identical settings as for Chinese, which is expected to lead to suboptimal performance on the evaluation data. Nevertheless, we measure 27.09\% accuracy on the PDTB test set (surpassing the majority class baseline of 22.01\%), which shows that the model has potential to generalize across implicit discourse relations in a different language.

\begin{figure*}[h!t]
\includegraphics[width=2.0\columnwidth]{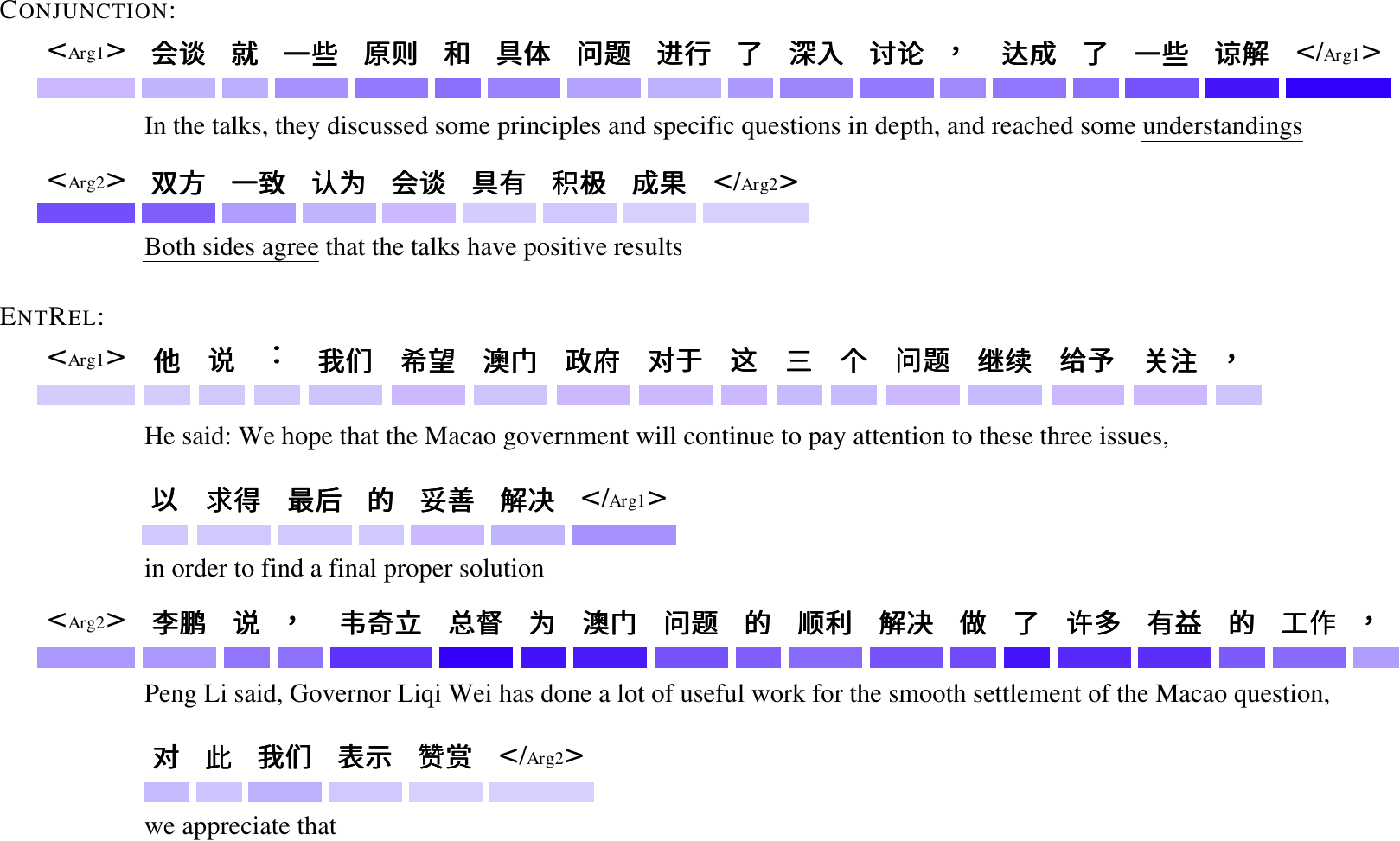}%
\caption{Visualization of attention weights for Chinese characters with high (dark blue) and low (light blue) intensities. The underlined English phrases are semantically structure-shared by the two arguments.}
\label{attentionweights}
\end{figure*}

\smallskip
\noindent \textbf{Visualizing Attention Weights:} Finally, in Figure \ref{attentionweights}, we illustrate the learned attention weights which pinpoint important subcomponents within a given implicit discourse relation. For the implicit \mbox{\textsc{Conjunction}} relation the weights indicate a peak on the transition between the argument boundary, establishing a connection between the semantically related terms \emph{understandings--agree}. Most \textsc{EntRel}s show an opposite trend: here second arguments exhibit larger intensities than \emph{Arg1}, as most entity relations follow the characteristic writing style of newspapers by adding additional information by reference to the same entity.








\section{Summary \& Outlook}

In this work, we have presented the first attention-based recurrent neural sense labeler specifically developed for Chinese implicit discourse relations. Its ability to model discourse units sequentially and jointly has been shown to be highly beneficial, both in terms of state-of-the-art performance on the CDTB (outperforming word order-agnostic feedforward approaches), and also in terms of insightful observations into the inner workings of the model through its attention mechanism. The architecture is structurally simple, benefits from partial argument sampling, and can be easily adapted to similar relation recognition tasks. In future work, we intend to extend our approach to different languages and domains, e.g., to the recent data sets on narrative story understanding or question answering \cite{mostafazadeh-EtAl:2016:N16-1,DBLP:conf/asru/FengXGWZ15}.
We believe that recurrent modeling of implicit discourse information can be a driving force in successfully handling such complex semantic processing tasks.\footnote{The code involved in this study is publicly available at \url{http://www.acoli.informatik.uni-frankfurt.de/resources/}.}

\section*{Acknowledgments}

The authors would like to thank Ayah Zirikly, Philip Schulz and Wei Ding for their very helpful suggestions on an early draft version of the paper, and also thank the anonymous reviewers for their valuable feedback and insightful comments. 
 We are grateful to Farrokh Mehryary for technical support with the attention layer implementation. Computational resources were provided by CSC -- IT Centre for Science, Finland, and Arcada University of Applied Sciences, Helsinki, Finland. Our research at Goethe University Frankfurt was supported by the project `Linked Open Dictionaries (LiODi, 2015-2020)', funded by the German Ministry for Education and Research (BMBF).

\newpage

\bibliography{references}
\bibliographystyle{acl_natbib}



\end{document}